\documentclass[runningheads]{llncs}
\usepackage[T1]{fontenc}
\usepackage[table]{xcolor} % For table line colors
\usepackage{float}
\usepackage[hidelinks]{hyperref}
\usepackage{hyperref}
\usepackage{graphicx}
\usepackage{booktabs} % For formal tables
\usepackage{makecell}
\usepackage{multirow}
\usepackage{subcaption}
\usepackage{booktabs} % For prettier horizontal lines
\usepackage{amsmath} % For math symbols
\usepackage{rotating}

\usepackage{amsmath,amsfonts,bm}

\def\eqref#1{equation~\ref{#1}}
\def\1{\bm{1}}

\DeclareMathAlphabet{\mathsfit}{\encodingdefault}{\sfdefault}{m}{sl}
\SetMathAlphabet{\mathsfit}{bold}{\encodingdefault}{\sfdefault}{bx}{n}

\newcommand{\bfI}{{\bf I}}

\newcommand{\bfK}{{\bf K}}

\newcommand{\bfW}{{\bf W}}

\newcommand{\bfb}{{\bf b}}

\newcommand{\bfx}{{\bf x}}

\newcommand{\bfu}{{\bf u}}

\newcommand{\hf}{{\frac 12}}

\newcommand{\bftheta}{{\boldsymbol \theta}}

\newcommand{\bfkappa}{{\boldsymbol \kappa}}

\begin{document}
\title{Beyond Conventional Parametric Modeling: Data-Driven Framework for Estimation and Prediction of Time Activity Curves in Dynamic PET Imaging}

% \author{Anonymous}
% % For review purposes, do not include any author information in the author section
% % below the title. This includes names, ORCID IDs, affiliations, email addresses, and URLs. 
% % Use asterisks (i.e. ***) or “anonymous”. DO NOT remove the author section to gain
% % extra writing space.

% \authorrunning{***}
% % % If there are more than two authors, '***' is used.

% \institute{Anonymous Organization}
% % Use asterisks (i.e. ***) or “anonymous” for affiliations, emails, and URLs.

\author{Niloufar Zakariaei\inst{1,2} \and Arman Rahmim\inst{1,2,3,4} \and Eldad Haber\inst{5}}

\authorrunning{Zakariaei et al.}

\institute{
School of Biomedical Engineering, University of British Columbia \and
Department of Integrative Oncology, BC Cancer Research Institute \and
Department of Radiology, University of British Columbia \and
Department of Physics, University of British Columbia\and
Department of Department of Earth, Ocean and Atmospheric Sciences\\
\email{nilouzk@student.ubc.ca}, \email{Arman.rahmim@ubc.ca}, \email{ehaber@eoas.ubc.ca}
}

\maketitle              % typeset the header of the contribution
\begin{abstract}
Dynamic Positron Emission Tomography (dPET) imaging and Time-Activity Curve (TAC) analyses are essential for understanding and quantifying the biodistribution of radiopharmaceuticals over time and space. Traditional compartmental modeling, while foundational, commonly struggles to fully capture the complexities of biological systems, including non-linear dynamics and variability. This study introduces an innovative data-driven neural network-based framework, inspired by Reaction Diffusion  systems, designed to address these limitations. Our approach, which adaptively fits TACs from dPET, enables the direct calibration of diffusion coefficients and reaction terms from observed data, offering significant improvements in predictive accuracy and robustness over traditional methods, especially in complex biological scenarios. By more accurately modeling the spatio-temporal dynamics of radiopharmaceuticals, our method advances modeling of pharmacokinetic and pharmacodynamic processes, enabling new possibilities in quantitative nuclear medicine.

\end{abstract}
\section{Introduction}

Positron Emission Tomography (PET) is a medical imaging technique that uses radioactive pharmaceuticals to visualize and measure body physiological processes, widely applied in oncology, neurology, and cardiology \cite{Trotter2023,Jadvar2005,9201038}. Dynamic PET, through sequential imaging, allows for the detailed visualization and quantification of radiopharmaceutical distribution within the body over time, providing valuable insights into physiological processes such as blood flow, glucose metabolism, and receptor binding \cite{Rahmim2019Dynamic}.

The processing of dynamic PET data involves image reconstruction followed by generation and analysis of Time-Activity Curves (TACs). TACs are generated by plotting the radioactivity concentration over time in specific regions of interest (ROI) \cite{8735911}. Subsequently, pharmacokinetic modeling is used to estimate physiological parameters related to the radiopharmaceutical's behavior. Commonly, {\em parametric models} are used. Such models are based on compartmental modeling, incorporating different physiological compartments and their interconnections. These models are dubbed as parametric models because they contain a small number of parameters (typically less than 10) that govern the shape of the TAC \cite{dimitrakopoulou2021kinetic,10.1007/978-3-031-43907-0_28}.

In principle, a reasonable parametric pharmacokinetic model should fit a TAC and be flexible enough to accommodate different TACs from different regions and patients. In particular,  Two-Tissue Compartment Model (2TCM) and  Three-Tissue Compartment Model (3TCM) are common in practice \cite{dimitrakopoulou2021kinetic,9201038}. Since such models contain only a few parameters, they are straightforward to use in order to explain dynamics. However, these models rely on strong assumptions that may not hold under complex biological conditions, potentially limiting their applicability in realistic settings \cite{keeling2008,diekmann2012}. For instance, these models assume it is possible to homogenize each compartment to a constant rate of exchange between compartments, neglecting the variability and non-linear dynamics characteristic of biological systems. Furthermore, these models do not commonly consider the impact of physiological changes in the course of imaging or the influence of underlying pathologies that may alter radiopharmaceutical kinetics and distribution, which can lead to inaccuracies in PET signal interpretation \cite{anderson1991,jama2020,morris2004kinetic}. There are also issues of patient and/or organ movements over time that may not be fully compensated even with motion correction methods, impacting kinetic models \cite{kotasidis2014advanced}. Finally, research in fields such as emergence \cite{cucker2007mathematics} suggests that combining even simple properties (like the decay at each cell) can yield a much more complex behavior of the homogenized system.

\begin{figure}[h]
    \centering
    \begin{minipage}[c]{0.45\linewidth}
    \begin{tabular}{cccc}
    \includegraphics[width=1.20cm]{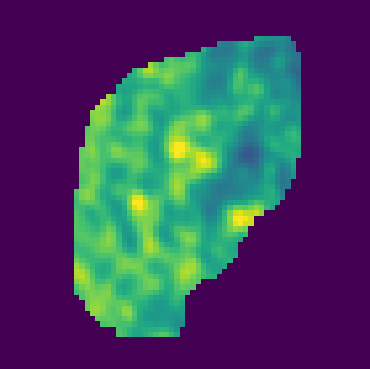} &
    \includegraphics[width=1.20cm]{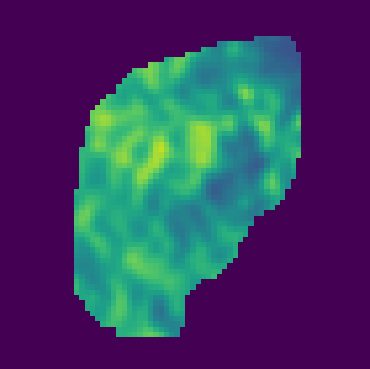} &
    \includegraphics[width=1.20cm]{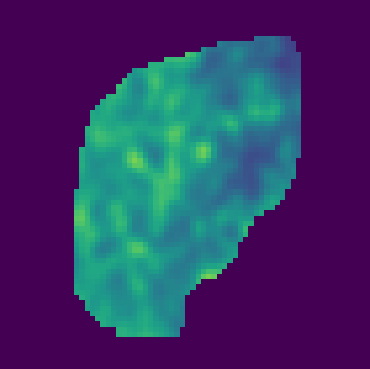} &
    \includegraphics[width=1.20cm]{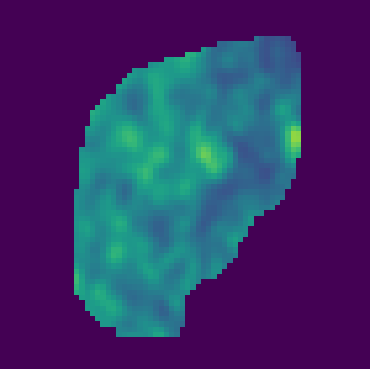} \\
    $t_1$ &  $t_2$ & $t_3$ & $t_4$ \\
    \includegraphics[width=1.20cm]{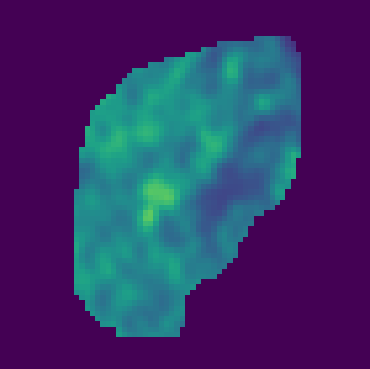} &
    \includegraphics[width=1.20cm]{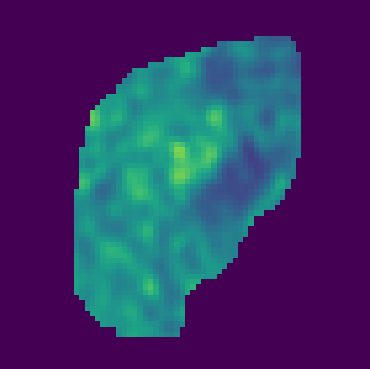} &
    \includegraphics[width=1.20cm]{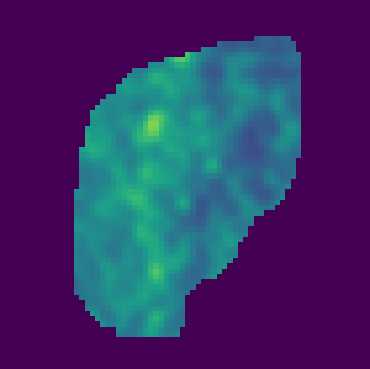} &
    \includegraphics[width=1.20cm]{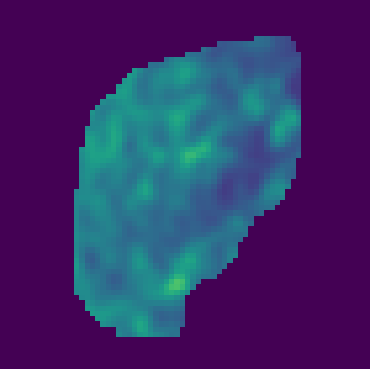} \\
    $t_5$ &  $t_6$ & $t_7$ & $t_8$ \\
    \end{tabular}
    \newline
    \begin{tabular}{c} % This centers the colorbar
    \includegraphics[width=0.9\textwidth]{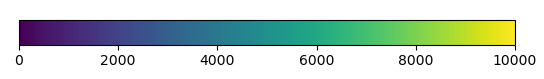} % Adjust the width as needed
    \end{tabular}
    \caption{Dynamic liver images captured at eight distinct time points.}
    \label{fig:fig1}
    \end{minipage}
    \hfill
\begin{minipage}[c]{0.45\linewidth}
\centering
    \includegraphics[width=4.5cm]{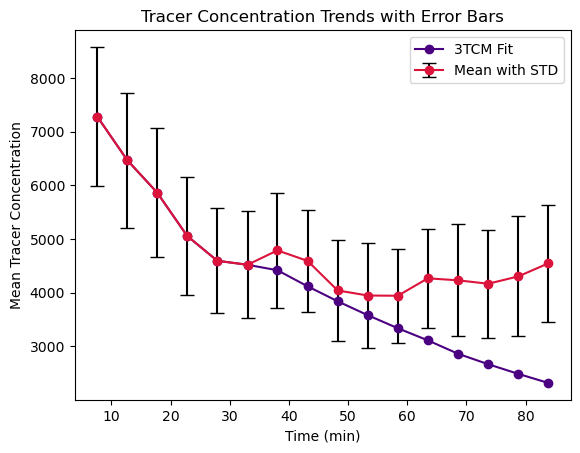}
        \caption{The average of the TAC curve over the liver, along with its standard deviation. Data is fitted using a three-tissue compartment model (3TCM) based on the early 6 time frames.}
    \label{fig:fig2}
    \end{minipage}
\end{figure}

These limitations highlight the need to extend or refine compartmental models to better capture the complexity of biological systems and enhance PET analysis reliability. 
% A clear example of the potential mismatch between model predictions and actual observations is illustrated in Figure 1, showcasing 8 liver slices at different time points, which may reveal discrepancies due to the oversimplified assumptions of the models.
A motivating example for this is demonstrated in Figure~\ref{fig:fig1} where 1 slice of a liver are given at different times.

While some decay is evident initially, it is challenging to justify a specific parametric form. Moreover, averaging across the entire organ yields a curve (see Figure~\ref{fig:fig2}) that does not exhibit typical compartment behavior of single- or multi-exponential biological decay. Observation of the data reveals spatial patterns that are that are not resolved by averaging activity over the organ. These patterns suggest variability in liver function, which could be insightful for therapeutic treatments.

{\bf The goal of this paper} is to propose a {\bf data-driven} methodology for the estimation of TACs, that vary both temporally and spatially. Instead of using a particular parametric form for a TAC, we utilize a carefully designed neural network architecture that aligns with the physical behavior of the phenomena we observe, calibrating its parameters based on the data at hand. Neural network models, far more complex than simple compartmental models, are justified only by their enhanced predictive power. In this work, we show that indeed, these models outperform simple compartmental models, at least for the datasets we have tested.

Neural network architectures that mimic physical systems are now common in many fields of science and engineering \cite{HaberRuthotto2017,jin2017deep,RuthottoHaber2018}. Such architectures, based on Partial Differential Equations in high dimensions, can be tailored to different physical phenomena, contrasting with compartmental models based on fitting very few parameters in an Ordinary Differential Equations. As highlighted in the literature \cite{HaberRuthotto2017}, this approach allows for more tailored and nuanced simulations of physical systems, offering potential advantages over compartmental models in terms of both precision and applicability. Here, we choose an architecture based on a Reaction Diffusion system. The data is used to learn the diffusion coefficients and the reaction term. We show that such a network can fit spatio-temporal patterns observed in the data, obtaining stronger predictive power than classical parametric models. These models can be effectively trained, uncovering new patterns in the data that simple parametric models cannot resolve. Moreover, by using a physically motivated neural network architecture, our network can be seen as an extension of a compartment model into high dimensions.

\section{Deriving a reaction diffusion  neural network for the modeling of TACs}

Reaction Diffusion systems have been used extensively in biology \cite{murray2,schnackenberg,kondo2010reaction}, from modeling the propagation of electromagnetic waves in the heart \cite{panfilov2019reaction} to patterns generated on butterfly wings \cite{kondo2010reaction}. The equations describe the interaction between a number of species (that can be chemicals or different populations) and their spatial dynamics. 

Hinted by its name, the equation comprises two parts: reaction and diffusion. The reaction term is local, meaning it is pointwise. Compartment models can, in fact, be considered as local reaction terms. The diffusion represents the spatial dynamics. Typically, different species have different diffusion coefficients. The interaction between the diffusion and reaction terms is the cause of pattern formation (see the classical work by Turing \cite{turing1990chemical} and references within).
The reaction-diffusion equation can be written as
\begin{eqnarray}
\label{rd}
{\frac {\partial \bfu}{\partial t}} =
\bfkappa \Delta \bfu + R(\bfu, t; \bftheta)
\end{eqnarray}
with appropriate initial and boundary conditions. Here, $\bfu= [\bfu_1, \ldots, \bfu_c]$ is a vector representing $c$ different species (or in the context of deep networks, channels), and the coefficients $\bfkappa=[\bfkappa_1, \ldots, \bfkappa_s]$ are the diffusion coefficients for each species. The reaction term $R(\bfu, t; \bftheta)$ couples the different species through nonlinear interaction. Finally, the parameters $\bftheta$ are trainable parameters in the reaction term.% from data.

While reaction diffusion systems have been applied to physical and biological systems for a long time, recent developments in neural network technology have demonstrated that it is possible to derive a neural network interpreted as a reaction diffusion system in high dimensions \cite{RuthottoHaber2018}. In the context of dynamic PET, the image under consideration can be thought of as a weighted sum of different species. In the 2 or 3 compartment model, only 2 or 3 species are used; however, with a deep network, one can employ an arbitrarily large number of species and learn the reaction term, that is, the interaction between them.

Let $I(t, \mathbf{x})$ represent the PET image we aim to model, where both $\mathbf{x}$ and $t$ are discretizations of space and time. Initially, the network embeds $I(\mathbf{x}, t)$ into a higher-dimensional state using a so-called Multilayer Preceptron (MLP) \cite{KrizhevskySutskeverHinton2012}. Let $\mathbf{u}(\mathbf{x}, t) = [\mathbf{u}_1(\mathbf{x}, t), \ldots, \mathbf{u}_c(\mathbf{x}, t)]$ be the embedded state. We assume $\mathbf{u}$ adheres to the reaction diffusion \eqref{rd} and discretize it in space-time. A common approach for the discretization of such a system is the Implicit-Explicit method (IMEX) (see \cite{ascher,Ruuth2,ruuthphd,ascherBook} and references within). Let $\mathbf{A}$ be a discretization of the negative Laplacian. Then the discretization of the system reads
\begin{eqnarray}
    \label{rddi}
  \tilde{\mathbf{u}}_{k+1} -\mathbf{u}_k = 
    -h \kappa \mathbf{A} \mathbf{u}_{k+1} 
      \label{rdde} \quad {\rm and}
\quad \mathbf{u}_{k+1} -   \tilde{\mathbf{u}}_{k+1} = h R(\tilde{\mathbf{u}}_{k+1}, t; \boldsymbol{\theta}) 
\end{eqnarray}
where $h$ is the time step.
The first equation is implicit and requires the solution of the system 
\begin{eqnarray}
\label{lap}
   (\mathbf{I} + h \kappa \mathbf{A})\tilde{\mathbf{u}}_{k+1} =\mathbf{u}_k  
\end{eqnarray}
The solution of a linear system is, in general, slow. However, since the image is defined on a regular grid, the inversion of the method can be efficiently achieved in order $n\log(n)$ (where $n$ is the number of pixels) by using a cosine transform \cite{nagyHansenBook}.
The reaction term \eqref{rdde} is modeled by a two-layer MLP with the form
\begin{eqnarray}
\label{mlp}
   R(\mathbf{u}, t) = \mathbf{K}_2\sigma(\mathbf{K}_1\mathbf{u} + \mathbf{t}_e).
\end{eqnarray}
Here, $\mathbf{K}_1$ and $\mathbf{K}_2$ are so-called $1\times 1$ convolutions; i.e., they only mix the channels of $\mathbf{u}$. The function $\sigma$ is an activation function (here we have used the silu activation) and $\mathbf{t}_e$ is an embedding of the time. Here, similar to diffusion-based methods \cite{song2019generative}, we have used an MLP 
\begin{eqnarray}
    \label{time_embedding}
    \mathbf{t}_e = \mathbf{W} \sigma( t \mathbf{b}) 
\end{eqnarray}
where $t$ is the scalar input time and $\mathbf{b}$ and $\mathbf{W}$ are learnable parameters.

\begin{figure}
    \centering
    \includegraphics[width=8cm]{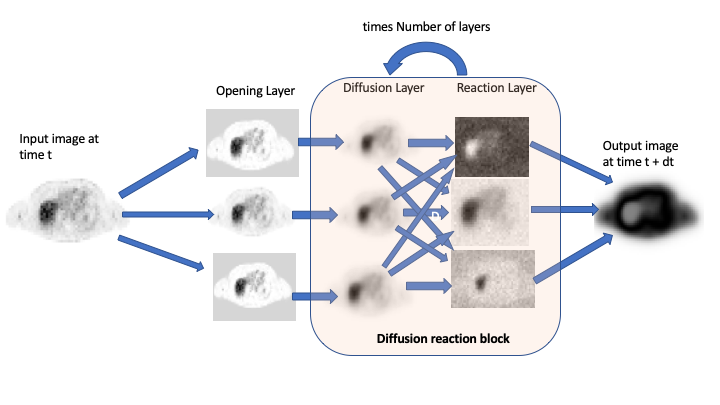}
    \caption{The reaction diffusion network architecture: An opening layer takes the input image into a higher embedded dimension, where a reaction diffusion network with learned diffusivity and reaction operates. In the example above, 3 channels are utilized to open the image.}

    \label{fig:net}
\end{figure}

The output of the network is an image: therefore, the network has one final so-called closing layer that compresses the output of the reaction diffusion discretization into a single channel. A sketch of our network is plotted in Figure~\ref{fig:net}.
Given the time-dependent data, our network learns the MLP that embeds the image, the diffusion coefficients $\bfkappa$, the convolution matrices $\bfK_1$, $\bfK_2$, and the time embeddings  $\bfb$ and $\bfW$. Finally, the network learns the closing MLP layer.

Note that similar to Long short-term memory (LSTM's)  our network shares parameters between different time steps \cite{gconvlstm} and also unlike LSTM our network uses time embedding, which makes the function explicitly depend on time. We observed that this allows us to better predict the time behaviour of the system. 

\section{Training the network and Dataset}
\label{sec:training}
%\subsection{Training}
As stated in the introduction, a neural network model can be justified to use if it yields better predictions. Our training process is geared to demonstrate that. 
Let $I(t_j,\bfx), j=1,\ldots,N$ be a set of images obtained from a single patient. For shorthand we define $\bfI_j=I(t_j,\bfx)$.
To this end, we have divided the data set  into two groups: a training group $\bfI_1, \ldots, \bfI_s$ and a validation set $\bfI_{s+1}, \ldots,\bfI_{N}$. Our goal is to train on the training images, in times $t_1, \ldots, t_s$ and for the network to predict the later images that are in the validation set.
In the training, we assume to have an image $\bfI_j$ and we attempt to predict $\bfI_{j+1}$, that is, 
\begin{eqnarray}
    \label{pred}
    \bfI^{\rm pred}_{j+1} = f(\bfI_j, \bftheta)
\end{eqnarray}
where $f(\cdot, \cdot)$ is the neural network described above and $\bftheta$ are the neural network parameters. 
To calibrate the parameters we minimize the standard Mean Squared Error (MSE) loss
%\begin{eqnarray}
%    \label{min}
 $    \hf \sum_j \|\bfI^{\rm pred}_{j+1}(\bftheta) - \bfI_{j+1} \|^2$.
%\end{eqnarray}
The training method is commonly used for LSTM networks \cite{gclstm} and has also been recently proposed for training reaction diffusion networks in the context of graph neural networks \cite{eliasof2023adr}. 

For the minimization process, we utilize the Adam optimizer \cite{kingma2014adam}, employing gradient clipping to ensuring the stability. To test the predictability of our model, our network is trained only on the first $s$ time steps, and then it is used to predict the subsequent activity. As mentioned in the introduction, the justification for using a complex model such as a neural network is its ability to predict the TAC beyond its training point, which is crucial for validation. As demonstrated in the following section, our model successfully achieves this objective.

Our research employs a dataset from 7 male patients undergoing [\textsuperscript{18}F]DCFPyL imaging for prostate cancer. The collected data reveal an average age of $70.29 \pm 1.89$ years, a mean weight of $93.71 \pm 18.20$ kg and an average administered dose of $7.08 \pm 1.31$ mCi.

\section{Numerical Results and Evaluation}

\begin{figure}[h!]
    \centering
    \begin{tabular}{cccc}
    % \multicolumn{4}{c}{\textbf{Ground Truth}} \\
    \multicolumn{4}{c}{\scalebox{0.8}{\textbf{Ground Truth}}}\\
    \end{tabular}
    
    \begin{minipage}{0.20\textwidth}
        \includegraphics[width=\linewidth]{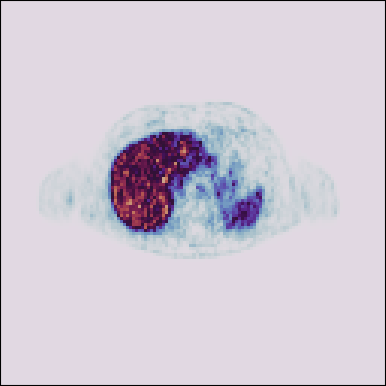}
        \centering
        \scalebox{0.8}{$t= 68 \, \text{min}$}
    \end{minipage}%
    %\hspace{0 mm}
    \begin{minipage}{0.20\textwidth}
        \includegraphics[width=\linewidth]{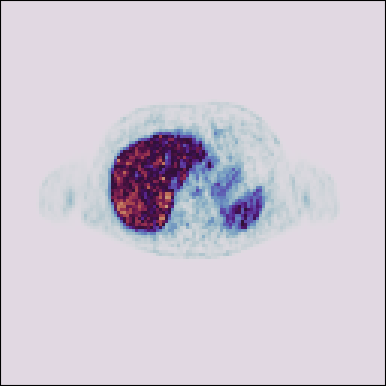}
        \centering
        \scalebox{0.8}{$t= 73 \, \text{min}$}
    \end{minipage}%
    %\hspace{0 mm}
    \begin{minipage}{0.20\textwidth}
        \includegraphics[width=\linewidth]{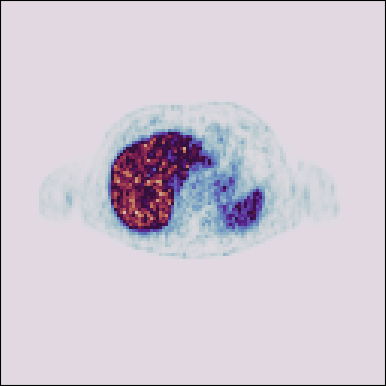}
        \centering
        \scalebox{0.8}{$t= 78 \, \text{min}$}
    \end{minipage}%
    %\hspace{0mm}
    \begin{minipage}{0.20\textwidth}
        \includegraphics[width=\linewidth]{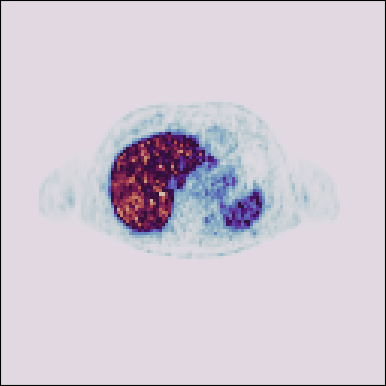}
        \centering
        \scalebox{0.8}{$t= 83 \, \text{min}$}
    \end{minipage}
    \vspace{1mm} % Adjust space between rows

    %\vspace{1mm} % Adjust space between rows
    
    \begin{tabular}{cccc}
    %\multicolumn{4}{c}{\textbf{3TCM}} \\
    \multicolumn{4}{c}{\scalebox{0.8}{\textbf{3TCM}}}\\
    \end{tabular}
    
    \begin{minipage}{0.20\textwidth}
        \includegraphics[width=\linewidth]{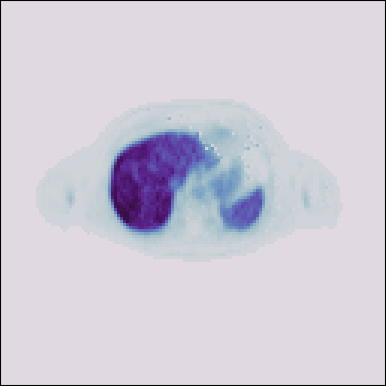}
        \centering
        %$t_5= 68 min$
        \scalebox{0.8}{$t= 68 \, \text{min}$}
    \end{minipage}%
    %\hspace{0mm}
    \begin{minipage}{0.20\textwidth}
        \includegraphics[width=\linewidth]{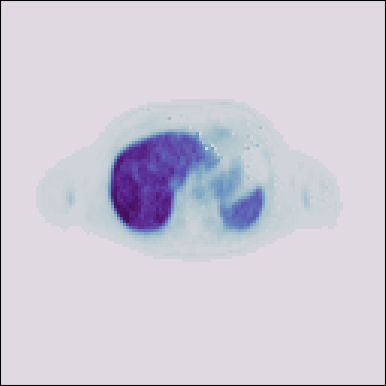}
        \centering
        %$t_6= 73 min$
        \scalebox{0.8}{$t= 73 \, \text{min}$}
    \end{minipage}%
    %\hspace{0mm}
    \begin{minipage}{0.20\textwidth}
        \includegraphics[width=\linewidth]{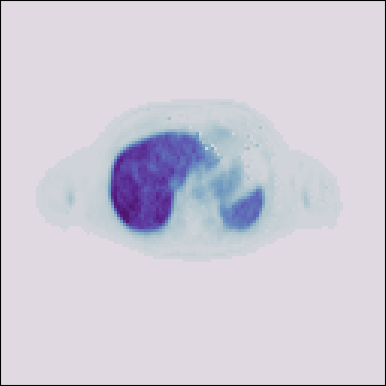}
        \centering
        %$t_7= 78 min$
        \scalebox{0.8}{$t= 78 \, \text{min}$}
    \end{minipage}%
    %\hspace{0mm}
    \begin{minipage}{0.20\textwidth}
        \includegraphics[width=\linewidth]{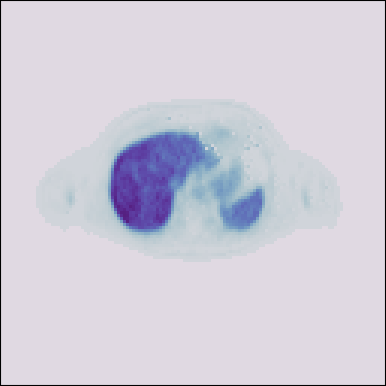}
        \centering
        %$t_8= 83 min$
        \scalebox{0.8}{$t= 83 \, \text{min}$}
    \end{minipage}
    \vspace{1mm}

        \begin{tabular}{cccc}
    %\multicolumn{4}{c}{\textbf{Reaction Diffusion}} \\
    %\multicolumn{4}{c}{{\small \textbf{Reaction Diffusion}}} \\
    \multicolumn{4}{c}{\scalebox{0.8}{\textbf{Proposed Model}}}\\
    \end{tabular}
    
    \begin{minipage}{0.20\textwidth}
        \includegraphics[width=\linewidth]{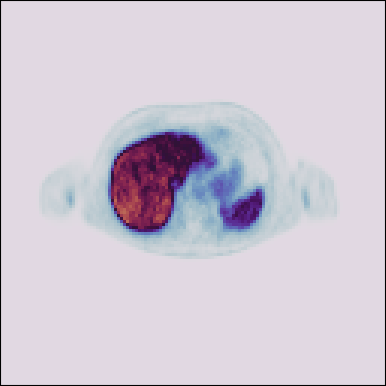}
        \centering
        %$t_5= 68 min$
        \scalebox{0.8}{$t= 68 \, \text{min}$}
    \end{minipage}%
    %\hspace{0mm}
    \begin{minipage}{0.20\textwidth}
        \includegraphics[width=\linewidth]{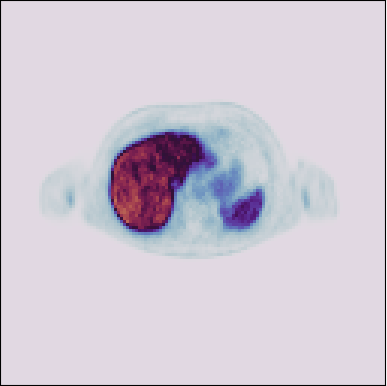}
        \centering
        %$t_6= 73 min$
        \scalebox{0.8}{$t= 73 \, \text{min}$}
    \end{minipage}%
    %\hspace{0mm}
    \begin{minipage}{0.20\textwidth}
        \includegraphics[width=\linewidth]{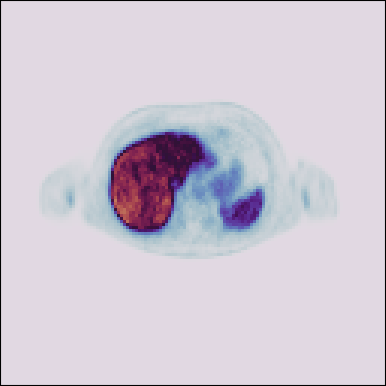}
        \centering
        %$t_7= 78 min$
        \scalebox{0.8}{$t= 78 \, \text{min}$}
    \end{minipage}%
    %\hspace{0mm}
    \begin{minipage}{0.20\textwidth}
        \includegraphics[width=\linewidth]{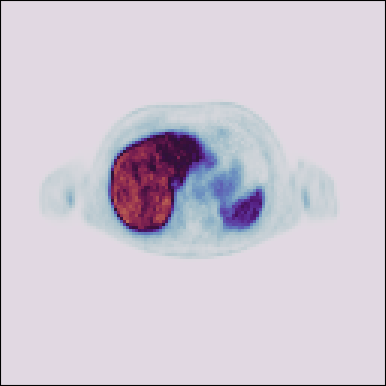}
        \centering
        %$t_8= 83 min$
        \scalebox{0.8}{$t= 83 \, \text{min}$}
    \end{minipage}
   
    \vspace{2mm} % Adjust space before the colorbar
    \includegraphics[width=0.65\textwidth]{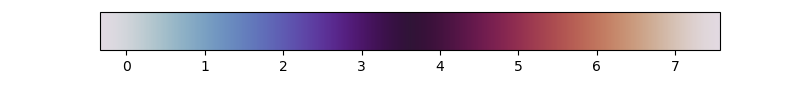} % Adjust the width as needed
    \caption{A selected slice of one of the patients, showcasing the latest time points predicted by our proposed model versus the 3TCM and the original images.}
    \label{fig: comparison}
\end{figure}

As outlined in Section~\ref{sec:training}, we used the early time frames (here the first 11 frames) of our dataset to train our network to predict the latest frames (last 4 frames). Similarly, we applied this method to the conventional 3TCM using non-linear least squares analysis. Finally, we compared the reaction diffusion network's performance against the conventional 3TCM, benchmarking both against ground truth. Figure \ref{fig: comparison} clearly illustrates that the reaction diffusion network aligns more closely with the actual data than the 3TCM in the latter four time frames. 

\begin{figure}[h!]
  \centering
  % Patient 1
  \begin{subfigure}{.28\textwidth}
    \includegraphics[width=\linewidth]{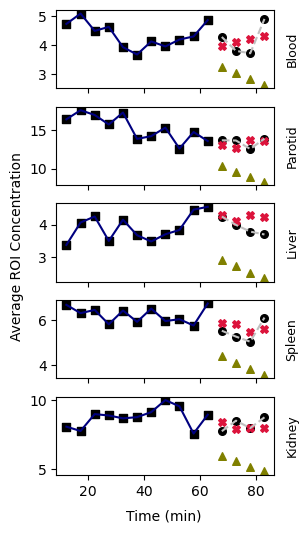}
  \end{subfigure}
  \hfill
  % Patient 2
  \begin{subfigure}{.28\textwidth}
    \includegraphics[width=\linewidth]{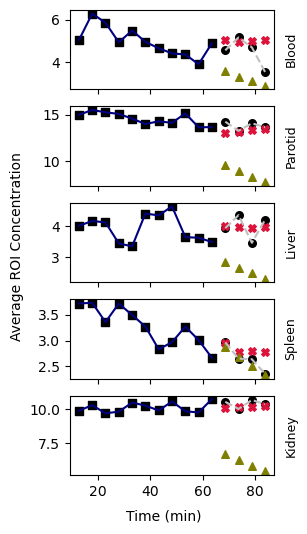}
  \end{subfigure}
  \hfill
  % Patient 3
  \begin{subfigure}{.28\textwidth}
    \includegraphics[width=\linewidth]{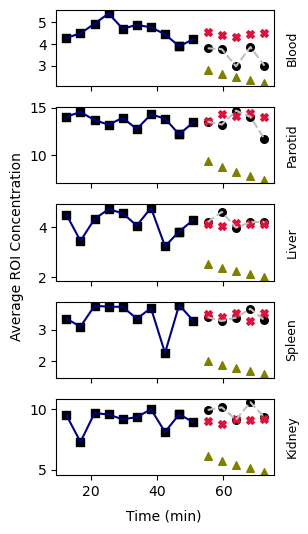}
  \end{subfigure}
  
  % Colorbar - spanning the width of three columns
  \begin{subfigure}{0.80\textwidth}
    \includegraphics[width=\linewidth]{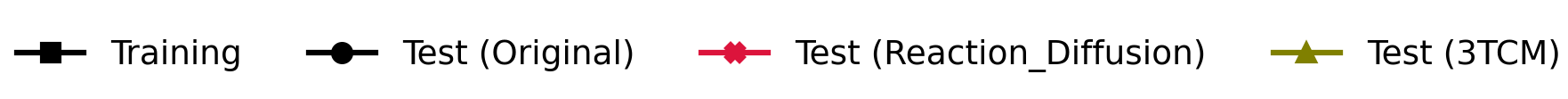}
    %\caption{Colorbar} % uncomment if a caption is needed for the colorbar
  \end{subfigure}
  
  \caption{Predicted vs. actual [\textsuperscript{18}F]DCFPyL concentrations over time in 3 patients. Training data are shown with solid markers, test data with hollow markers. Predictions from the reaction diffusion model and traditional 3TCM are compared with original test. Trend lines connect data points, showing our model's enhanced accuracy in capturing distribution dynamics.}
  \label{fig: TACs}
\end{figure}

The superiority of our model is further evidenced in Figure ~\ref{fig: TACs}, which provides a detailed examination of small, defined regions within each organ (15-30 pixels), as detailed in Table ~\ref{table:RD} and Table ~\ref{table:3TCM}. Here, we show the result for 3 patients as an example. The neural network's predicted TACs are shown to more accurately reflect the ground truth compared to those produced by 3TCM, which is less accurate when the data doesn't follow a decaying pattern.

The efficacy of our model is quantitatively substantiated in Table ~\ref{table:RD} and Table ~\ref{table:3TCM} , which uses the MSE metric, calculated over the test time frames, highlighting our model's predictive strength for future time points. For a robust analysis, five characteristic slices of the targeted organ from each patient were analyzed, aligning with the test time frames.

\newcolumntype{C}[1]{>{\centering\arraybackslash}m{#1}}

\begin{table}[h!]
  \centering

  % First minipage for the first table
  \begin{minipage}[t]{0.48\textwidth}
    \centering
    \caption{Proposed Model}
    \label{table:RD}
    \begin{tabular}{|C{0.9 cm}|C{0.9 cm}|C{0.9 cm}|C{0.9 cm}|C{0.9 cm}|C{0.9 cm}|}
      \hline
      \rotatebox{60}{\textbf{Patient}} & \rotatebox{60}{\textbf{Parotid}} & \rotatebox{60}{\textbf{Blood}} & \rotatebox{60}{\textbf{Liver}} & \rotatebox{60}{\textbf{Spleen}} & \rotatebox{60}{\textbf{Kidney}} \\
      \hline
      1 & 0.685  & 0.195 & 0.134& 0.207 & 0.343\\
      
      \hline
      2 & 1.974 & 0.441 & 0.115 &0.062 & 0.212 \\
      
      \hline
      3 & 0.553 & 0.665 & 0.105& 0.054 &0.121 \\
     
      \hline
      4 & 2.752 & 0.424 & 0.505 & 0.151 &1.037 \\
      
      \hline
      5 & 0.657 & 1.045 & 0.687& 0.351 & 1.099\\
      
      \hline
      6 & 1.404 &1.085 & 0.072& 0.051& 0.942 \\
      
      \hline
      7 & 0.750 & 4.167&0.250 & 0.189& 0.236\\
     
      \hline
    \end{tabular}
  \end{minipage}
  \hfill
  % Second minipage for the second table
  \begin{minipage}[t]{0.48\textwidth}
    \centering
    \caption{3TCM}
    \label{table:3TCM}
    \begin{tabular}{|C{0.9 cm}|C{0.9 cm}|C{0.9 cm}|C{0.9 cm}|C{0.9 cm}|C{0.9 cm}|}
      \hline
      \rotatebox{60}{\textbf{Patient}} & \rotatebox{60}{\textbf{Parotid}} & \rotatebox{60}{\textbf{Blood}} & \rotatebox{60}{\textbf{Liver}} & \rotatebox{60}{\textbf{Spleen}} & \rotatebox{60}{\textbf{Kidney}} \\
      \hline
      1 & 18.08 & 1.927 & 1.671 & 2.611 & 8.767 \\
      
      \hline
      2 & 55.27 & 1.199 & 4.701 & 13.59 & 17.71\\
      
      \hline
      3 & 27.13 & 1.822 & 2.103& 0.007& 19.05\\
      
      \hline
      4 & 44.63 & 1.651& 3.856 & 0.715 & 18.84\\
      
      \hline
      5 & 28.28 & 0.282 & 1.661& 0.059& 2.969\\
      
      \hline
      6 &27.53 & 1.076& 4.073& 2.694& 19.60\\
      
      \hline
      7 &23.36  &4.167 &3.498 &1.758 & 11.27\\
      
      \hline
    \end{tabular}
  \end{minipage}
  %\vspace{1mm}
  \caption*{Comparison across 5 organs in 7 patients shows the proposed model's predictive accuracy vs. 3TCM over all test time frames, using MSE metric.}
  
\end{table}

\section{Discussion and Conclusions}
In this paper, we propose the use of a reaction diffusion neural network to model TACs in Dynamic PET imaging. These models replace the commonly used two- and three-compartment models. Although more complex, they offer higher flexibility in fitting data, allowing for better prediction of non-trivial TACs. It is important to note that reaction diffusion neural networks differ from "generic" neural networks and tend to behave similarly to their continuous analogs as extensively used in chemistry and biology \cite{turing1990chemical}. As such, these models can be considered "biological", similar to other reaction diffusion models in biology. The results obtained by our networks suggest they are far more predictive compared to compartmental models. In fact, our predictions of TACs are superior to any parametric model known to us, suggesting that our model should be considered when a  model is needed to predict TAC's. This is particularly meaningful, given the development of complex radiopharmaceuticals (e.g., $[^{18}\text{F}]$DCFPyL and next-gen theranostics) with mechanisms less known than those of older radiopharmaceuticals \cite{crișan2022radiopharmaceuticals}.

One important question remains unanswered: why are reaction diffusion models superior to simpler models that are pointwise (i.e., models that do not include a spatial term)? We hypothesize that the diffusion predicted by the model does not represent only diffusion but also the blurring of the image due to unaccounted-for recovery, motion errors and partial volume effects that leads to blurring which is equivalent to diffusion. The reaction term used in our network is simple, involving only two layers with a single non-linearity. Approximation theory suggests that such an architecture is sufficient to model any continuous function \cite{wu2020comprehensive}. Thus, while we have presented a data-driven approach to parameter estimation for TACs in dynamic PET, our model’s superior predictive power can be both motivated and explainable physically.

\newpage
\bibliography{biblio}

\end{document}